\documentclass[11pt]{article}
\usepackage[letterpaper,margin=1in]{geometry}
\usepackage{amsmath,amssymb,amsthm}
\usepackage{graphicx}
\usepackage{booktabs}
\usepackage{cite}
\usepackage{xcolor}
\usepackage[hidelinks]{hyperref}
\usepackage{authblk}

\newtheorem{proposition}{Proposition}
\newtheorem{theorem}{Theorem}
\newtheorem{assumption}{Assumption}
\theoremstyle{remark}
\newtheorem{remark}{Remark}
\newcommand{\R}{\mathbb{R}}
\newcommand{\sym}{\operatorname{sym}}

\title{Locating Contact on Tendon-Driven Continuum Robots From Tendon Measurements Alone via Constrained Moving Horizon Estimation}
\author{Haitham El-Hussieny}
\affil{Egypt-Japan University of Science and Technology (E-JUST), New Borg El-Arab, Alexandria, Egypt}
\date{November 2026}

\begin{document}
\maketitle

\begin{abstract}
Tendon-driven continuum robots can reach through constrained spaces, but contact along their bodies is invisible to the sensors at their motors. We show that this is a structural limitation rather than a tuning problem: the map from contact forces to tendon measurements has a null space of dimension at least seven, so no estimator that leaves the contact support free can locate contact. When the obstacle geometry is known, however, non-penetration and complementarity make the contact force locally identifiable, and we prove that the resulting system with unknown contact inputs is locally exponentially strongly detectable. We present a constrained moving horizon estimator (MHE) that encodes these conditions and runs in real time as one quadratic program per sample, and we give a local, mode-wise robust stability result for it. In a finite-element benchmark of a two-section robot pressing and sliding on a wall, the estimator located contact to 0.9~mm and estimated contact force to 0.22~N, compared with 65~mm and 0.72~N for a tuned extended Kalman filter (EKF) with state projection. The estimator tolerated millimeter-level errors in the obstacle position and unmodeled wall friction, under which the projected EKF failed. Under randomized model mismatch over 40 trials, including unmodeled tendon friction, it remained better in force and location in every trial. Without contact, the MHE offered no advantage over the EKF, which confirms that its benefit comes from the contact constraints. This report extends a letter submitted to IEEE Control Systems Letters with complete proofs and additional experiments.
\end{abstract}

\section{Introduction}
Continuum robots could act as our hands in spaces that rigid instruments cannot reach, such as the anatomy traversed in minimally invasive surgery \cite{burgnerkahrs2015}. In these spaces the robot body touches its surroundings along its length, and knowing where and how hard it touches is needed both for safety and for control. However, sensing contact on a continuum robot is challenging, because force sensors are difficult to integrate into slender, flexible, and often sterilizable bodies.

Existing methods therefore infer contact from the robot's own deflection. Table~\ref{tab:related} summarizes the closest work. Xu and Simaan \cite{xu2010} estimated external wrenches from actuation forces, and Bajo and Simaan \cite{bajo2012} detected and localized contacts from kinematic deviations of multisegment robots. Aloi \emph{et al.} \cite{aloi2022} estimated distributed loads and their locations from shape measurements with a Cosserat model and an EKF, Ferguson \emph{et al.} \cite{ferguson2024} estimated shape and external loads jointly, Johnston \emph{et al.} \cite{johnston2025} used a modal momentum observer, and Prakash \emph{et al.} \cite{prakash2026} estimated multiple contact forces with a factor graph from strain and pose measurements. Learned methods have used distributed optical-fiber strain \cite{tavangarifard2026}. Each of these methods that localizes contact along the body relies on a shape signal. The only method we found that uses tendon measurements alone \cite{feng2021} estimates the force at the tip, not where contact occurs.

\begin{table}[t]
\centering
\caption{Contact estimation for continuum robots: sensing and estimated quantities.}
\label{tab:related}
\small
\begin{tabular}{@{}llll@{}}
\toprule
Work & Sensing & Estimates & Estimator \\
\midrule
Xu \& Simaan \cite{xu2010} & actuation forces, configuration & wrench & statics \\
Bajo \& Simaan \cite{bajo2012} & segment kinematics & contact, segment & thresholds \\
Aloi \emph{et al.} \cite{aloi2022} & 3D shape, tensions & loads and locations & EKF \\
Ferguson \emph{et al.} \cite{ferguson2024} & position or strain & shape and loads & batch estimation \\
Johnston \emph{et al.} \cite{johnston2025} & shape, motor current & wrench & momentum observer \\
Prakash \emph{et al.} \cite{prakash2026} & strain, tip pose, tensions & multiple contact forces & factor graph \\
Tavangarifard \emph{et al.} \cite{tavangarifard2026} & distributed fiber strain & contact, location, force & learned \\
Feng \emph{et al.} \cite{feng2021} & tendon tension and displacement & tip force & learned \\
This work & tendon tension and displacement & shape, force, location & constrained MHE \\
\bottomrule
\end{tabular}
\end{table}

Building on these results, there is a need for an estimator that locates and sizes contact along a tendon-driven robot using only the signals available at its motors. This report presents such an estimator, explains why it requires constraints that a Kalman filter cannot represent, and states when it is stable. Constrained moving horizon estimation (MHE) \cite{rao2003,rawlings2017} with contact complementarity has been used for legged robots \cite{kang2024}, where contact sensors provide the active contact set; here the active set must be inferred from geometry. The contributions of this work are:
\begin{enumerate}
\item \textbf{An identifiability and detectability analysis of contact from tendon measurements.} We prove that contact forces are not identifiable from tendon lengths without information on the contact support (Proposition~\ref{prop:nonid}), and that complementarity with a known obstacle restores local identifiability (Proposition~\ref{prop:id}) and local exponential strong detectability (Proposition~\ref{prop:det}). In contrast to prior observability discussions \cite{johnston2025}, which concern wrench components at a known contact, our analysis addresses where along the body contact occurs.
\item \textbf{A real-time constrained MHE with a stability result.} We encode non-penetration, non-negative forces, and complementarity in an MHE and solve it with a real-time iteration \cite{diehl2005,wynn2014}, replacing complementarity by a linearized penalty. We give a local, mode-wise robust stability result (Theorem~\ref{thm:stab}). In contrast to \cite{kang2024}, our estimator infers the active contact set from geometry.
\item \textbf{A finite-element benchmark against tuned Kalman filtering.} We compare the estimator with tuned EKFs, including state projection \cite{simon2010}, under nominal conditions, obstacle-position error, unmodeled wall friction, mid-body contact, randomized model mismatch, and, as a control, motion without contact.
\end{enumerate}

The remainder of this report is organized as follows. Section~\ref{sec:model} describes the robot, the reduced model, and the contact model. Section~\ref{sec:theory} presents the identifiability and detectability analysis. Section~\ref{sec:mhe} presents the estimator and its stability. Section~\ref{sec:results} reports the simulation study, and Section~\ref{sec:concl} concludes.

\section{Robot, Model, and Measurements}\label{sec:model}
This section describes the robot, the reduced model used by all model-based estimators, and the contact model.

\subsection{Robot and sensors}
We consider a two-section tendon-driven robot of length 0.6~m. Each section has three tendons spaced at 120$^\circ$ and offset 20~mm from the backbone, and the tendons of the second section are rotated by 60$^\circ$ relative to the first. Twenty-four spacer disks of radius 26~mm route the tendons. The only measurements are the six tendon length changes at the motors, with 0.1~mm noise, and the six tendon tensions $T\in\R^6$, with 0.5~N noise, both sampled at 20~Hz. The length measurements include the elastic elongation of the tendons, which we compensate with a tendon stiffness that is deliberately 10\% different from the true value. The tensions are commanded by a feedforward allocation that keeps all tendons above a 2~N pretension.

\subsection{Reduced model}
The estimators use a pseudo-rigid-body rod with $n_J=12$ joints placed at the midpoints of 12 equal segments, so that a constant-curvature arc is reproduced exactly at the joints. Each joint has two bending angles and one torsion angle, so the configuration is $q\in\R^{n_q}$ with $n_q=36$. We include torsion because a lateral contact at the tip twists the backbone by approximately 30$^\circ$ in our scenarios: a 1.85~N contact force acting at a lever of approximately 0.2~m against a torsional stiffness $GJ=0.39$~N\,m$^2$. Without torsion, every model-based estimator we tested failed in sliding contact, with shape errors of 9--12~mm instead of 2--3~mm.

Tendon-driven continuum robots move slowly relative to their elastic dynamics, so we model the motion as overdamped and write it with implicit Euler discretization as
\begin{equation}\label{eq:dyn}
\tfrac{d}{h}\,(q_{k+1}-q_k) = e(q_{k+1},\lambda_{k+1},T_k),
\end{equation}
\begin{equation}\label{eq:e}
e(q,\lambda,T) = -Kq-\nabla V(q)-J_l(q)^{\top} T-J_x(q)^{\top}\lambda,
\end{equation}
where $h=0.05$~s is the sample time; $d$ is a scalar damping coefficient; $K$ is the diagonal joint stiffness, with entries $EI/\ell$ for bending and $GJ/\ell$ for torsion, segment length $\ell=0.05$~m, and $GJ=EI/(1+\nu)$; $V$ is the gravitational potential of the disks; $l(q)\in\R^6$ are the tendon path lengths, computed as polylines through the disk holes, with Jacobian $J_l$; and $J_x$ is the Jacobian of the candidate contact points defined below. The tendon term follows from virtual work: a tendon under tension $T_i$ does work $-T_i\,\mathrm{d}l_i$. We chose implicit Euler because the elastic dynamics are stiff at $h=0.05$~s. We calibrated $d$ on one free-motion training run by matching the open-loop shape; the best value, $d=1.5$~N\,m\,s, agrees with $0.15\,EI/\ell$, the structural damping of the ground truth, and we use it unchanged throughout. The measurement model is
\begin{equation}\label{eq:y}
y_k = l(q_k)-l(0)+v_k,
\end{equation}
where $v_k$ is measurement noise with standard deviation $\sigma_y$.

\subsection{Contact model}
\begin{figure}[t]
\centering
\includegraphics[width=0.62\textwidth]{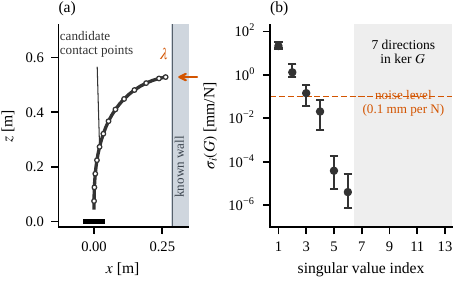}
\caption{Problem setting. (a) Finite-element ground truth during contact (run~1, side view), with the 13 candidate contact points and the true contact force $\lambda$ at the tip. (b) Singular values of the sensitivity $G$ from contact forces to tendon measurements (median and 10th--90th percentiles over 990 configurations). Seven directions lie in $\ker G$ and only approximately three exceed the noise level.}
\label{fig:setup}
\end{figure}

The obstacle is a plane at a known position $x_w$ along its normal. We place $n_c=13$ candidate contact points at the 12 joints and the tip, with normal forces $\lambda\in\R^{n_c}$ and gaps
\begin{equation}
g(q)=x_w-r-x(q),
\end{equation}
where $x(q)\in\R^{n_c}$ are the point coordinates along the obstacle normal and $r=26$~mm is the spacer-disk radius, so that a zero gap corresponds to a disk rim touching the plane. Figure~\ref{fig:setup}(a) shows the candidate points on the ground-truth robot during contact. Physically admissible contact satisfies
\begin{equation}\label{eq:lcp}
\lambda\ge 0,\qquad g(q)\ge 0,\qquad \lambda_i\,g_i(q)=0\quad (i=1,\dots,n_c).
\end{equation}
The last condition, i.e., \emph{complementarity}, states that force acts only where the robot touches the obstacle. Contact in the estimators is frictionless; Section~\ref{sec:rob} tests the effect of friction in the ground truth.

\section{Identifiability and Detectability}\label{sec:theory}
This section shows that tendon measurements cannot locate contact on their own and that complementarity supplies the missing information. We first analyze the quasi-static map from forces to measurements, and then the dynamic system with unknown contact inputs.

\subsection{Quasi-static identifiability}
Let $H(q)=-\partial e/\partial q$, the Hessian of the elastic and gravitational potential plus the tendon-curvature term. In all configurations analyzed below, the symmetric part of $H$ was positive definite (Section~\ref{sec:verify}). At an equilibrium $e(q,\lambda,T)=0$, the implicit function theorem gives $q=q(\lambda,T)$ locally, and the sensitivity of the measurements to the contact forces is
\begin{equation}\label{eq:G}
G(q)=\frac{\partial y}{\partial \lambda}=J_l(q)\,\frac{\partial q}{\partial\lambda}=-J_l(q)\,H(q)^{-1}J_x(q)^{\top}\in\R^{6\times n_c}.
\end{equation}

\begin{proposition}[Non-identifiability]\label{prop:nonid}
Since $\operatorname{rank}G\le 6<n_c$, the null space $\ker G$ has dimension at least $n_c-6=7$. For any admissible $\lambda$ and any $\delta\in\ker G$ with $\lambda+\delta$ admissible, the two force distributions produce the same measurements to first order. Therefore no estimator that leaves the support of $\lambda$ free can identify $\lambda$ locally from $y$.
\end{proposition}
\begin{proof}
$G$ has six rows, so $\operatorname{rank}G\le6$ and $\dim\ker G\ge n_c-6$. Local identifiability at $\lambda$ requires the measurement map to be locally injective, which requires $\ker G=\{0\}$; this is impossible when $n_c>6$.
\end{proof}

\begin{proposition}[Identifiability under complementarity]\label{prop:id}
Let $A=\{i: g_i(q)=0\}$ be the active contact set, assume strict complementarity ($\lambda_i>0$ for $i\in A$), and let $G_A$ denote the columns of $G$ indexed by $A$. If $G_A$ has full column rank, then under \eqref{eq:lcp} the forces $\lambda_A$ are locally identifiable from $y$, and any unbiased estimator satisfies the Cram\'er--Rao bound
\begin{equation}\label{eq:crlb}
\operatorname{Cov}(\hat\lambda_A)\succeq\sigma_y^2\,(G_A^{\top}G_A)^{-1},\qquad \operatorname{Var}(\mathbf 1^\top\hat\lambda_A)\ge\sigma_y^2\,\mathbf 1^\top(G_A^{\top}G_A)^{-1}\mathbf 1 .
\end{equation}
\end{proposition}
\begin{proof}
Under \eqref{eq:lcp}, $\lambda_i=0$ for $i\notin A$, so the unknown reduces to $\lambda_A\in\R^{|A|}$. Locally, $y\approx y_0+G_A\Delta\lambda_A+v$ with Gaussian $v$, whose Fisher information is $G_A^{\top}G_A/\sigma_y^2$. This matrix is nonsingular if and only if $G_A$ has full column rank, in which case the map $\Delta\lambda_A\mapsto G_A\Delta\lambda_A$ is injective and \eqref{eq:crlb} follows from the Cram\'er--Rao inequality.
\end{proof}

\begin{remark}[Where the active set comes from]
Proposition~\ref{prop:nonid} shows that the active set cannot be read from $G$. It comes from geometry: the shape $q$ is pinned by the measurements together with the model \eqref{eq:dyn}, and the gap constraint with complementarity selects the points that touch the known obstacle. Complementarity is a nonconvex, combinatorial condition, which a Kalman filter cannot represent. An EKF that augments $\lambda$ with random-walk states inherits Proposition~\ref{prop:nonid}: its covariance grows along $\ker G$, and projecting each estimate onto $\{\lambda\ge0,\,g(q)\ge0\}$ \cite{simon2010} removes only the infeasible part of the resulting drift.
\end{remark}

\begin{figure}[t]
\centering
\includegraphics[width=\textwidth]{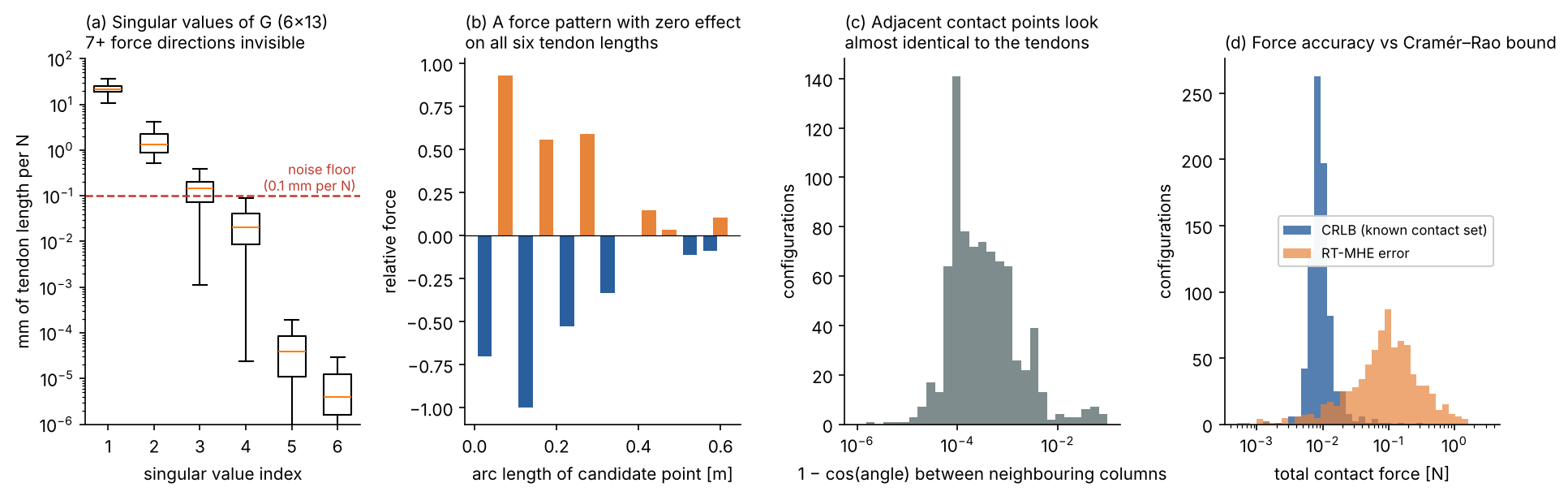}
\caption{Identifiability along five contact runs. (a) Singular values of $G$. (b) A force distribution in $\ker G$, which changes no tendon length. (c) Similarity of neighboring columns of $G$: adjacent contact points look almost identical to the tendons. (d) Cram\'er--Rao bound on the total contact force with a known active set, compared with the error of the real-time MHE; the gap of approximately 10$\times$ is due to model mismatch rather than sensor noise.}
\label{fig:ident}
\end{figure}

Figure~\ref{fig:ident} evaluates these quantities along five simulated contact runs (990 configurations, 805 in contact). The median singular values of $G$ were 21, 1.3, 0.14, 0.021, $4\times10^{-5}$, and $4\times10^{-6}$~mm/N, so with $\sigma_y=0.1$~mm only approximately three force directions are detectable at 1~N. Neighboring columns of $G$ were nearly parallel (median cosine 0.9997), so the measurements alone cannot distinguish adjacent contact points. The active set contained one point in 96\% and two points in 4\% of the contact configurations; $G_A$ had full column rank in all of them, with a median smallest singular value of 11~mm/N (minimum 0.05~mm/N). The Cram\'er--Rao bound on the total contact force had a median of 0.009~N (95th percentile 0.02~N).

\subsection{Strong detectability with unknown contact inputs}
For estimation over time, we linearize \eqref{eq:dyn}--\eqref{eq:y} along a reference trajectory $(q^\ast_k,\lambda^\ast_k)$. Writing $H_k=H(q^\ast_k)$, this gives
\begin{equation}\label{eq:lin}
\delta q_{k+1}=M_k\,\delta q_k+N_k\,\delta\lambda_{k+1},\qquad \delta y_k=C_k\,\delta q_k,
\end{equation}
with $M_k=(\tfrac{d}{h}I+H_k)^{-1}\tfrac{d}{h}$, $N_k=-(\tfrac{d}{h}I+H_k)^{-1}J_x(q^\ast_{k+1})^{\top}$, and $C_k=J_l(q^\ast_k)$. For an active set $A$, let $N_{A,k}$ denote the columns of $N_k$ indexed by $A$.

\begin{assumption}\label{ass:det}
For a fixed active set $A$ and a compact neighborhood of the reference trajectory:
\begin{enumerate}
\item[(S1)] $\sym(H_k)\succeq\mu I$ for some $\mu>0$;
\item[(S2)] $\sigma_{\min}(C_{k+1}N_{A,k})\ge\underline\sigma>0$;
\item[(S3)] the zero dynamics $\xi_{k+1}=P_kM_k\xi_k$, with $P_k=I-N_{A,k}(C_{k+1}N_{A,k})^{+}C_{k+1}$, are uniformly exponentially stable, i.e., $\|\prod_{j=k}^{k+L-1}P_jM_j\|\le\rho_L<1$ for some $L\ge1$ and all $k$.
\end{enumerate}
\end{assumption}

Condition (S1) states that elastic stiffness dominates, which makes the implicit step well posed. Condition (S2) is the discrete-time analogue of the rank condition $\operatorname{rank}(CB)=\operatorname{rank}(B)$ for unknown-input observers, and (S3) requires the states that produce zero output under a suitable unknown input to decay; together, (S2) and (S3) are the linear characterization of strong detectability \cite{hautus1983,guo2025}.

\begin{proposition}[Local exponential strong detectability]\label{prop:det}
Under Assumption~\ref{ass:det}, there exist $\varepsilon>0$, $c\ge1$, and $\hat\rho\in(0,1)$ such that any two solutions $(q,\lambda_A)$ and $(q',\lambda'_A)$ of the model \eqref{eq:dyn}--\eqref{eq:y} with additive process disturbances $w,w'$ that remain within $\varepsilon$ of the reference satisfy, for all $k\ge0$,
\begin{align}
|\Delta q_k|&\le c\,\hat\rho^{\,k}|\Delta q_0|+c\sum_{j=1}^{k}\hat\rho^{\,k-j}\big(|\Delta y_j|+|\Delta w_{j-1}|\big),\label{eq:det1}\\
|\Delta\lambda_{A,k}|&\le c\big(|\Delta q_{k-1}|+|\Delta y_k|+|\Delta w_{k-1}|\big),\label{eq:det2}
\end{align}
where $\Delta(\cdot)$ denotes the difference between the two solutions. That is, the system with unknown input $\lambda_A$ is locally exponentially strongly detectable in the sense of \cite{guo2025}.
\end{proposition}

\begin{proof}
Let $e_k=\Delta q_k$, $\delta_k=\Delta\lambda_{A,k}$, $\eta_k=\Delta y_k$, and $\omega_k=\Delta w_k$.

\emph{Step 1 (local expansion).} By (S1), $\tfrac{d}{h}I+H$ is invertible on the neighborhood, so by the implicit function theorem the step map defined by \eqref{eq:dyn} is $C^2$ and
\begin{align*}
e_{k+1}&=M_ke_k+N_{A,k}\delta_{k+1}+G_k\omega_k+r_k,\qquad |r_k|\le\ell\,(|e_k|+|\delta_{k+1}|)^2,\\
\eta_{k+1}&=C_{k+1}e_{k+1}+s_k,\qquad |s_k|\le\ell\,|e_{k+1}|^2,
\end{align*}
where $\ell$ bounds the second derivatives on the compact neighborhood and $\|G_k\|\le\bar g$.

\emph{Step 2 (input recovery).} Substituting the first line into the second gives $\eta_{k+1}=C_{k+1}M_ke_k+C_{k+1}N_{A,k}\delta_{k+1}+C_{k+1}G_k\omega_k+C_{k+1}r_k+s_k$. By (S2), $C_{k+1}N_{A,k}$ has a left inverse $(C_{k+1}N_{A,k})^{+}$ with norm at most $1/\underline\sigma$, so
\begin{equation*}
\delta_{k+1}=(C_{k+1}N_{A,k})^{+}\big[\eta_{k+1}-C_{k+1}M_ke_k-C_{k+1}G_k\omega_k-C_{k+1}r_k-s_k\big].
\end{equation*}
Since $M_k$, $C_k$, and $G_k$ are bounded and the remainders are quadratic, for $\varepsilon$ small enough this yields \eqref{eq:det2} with a suitable $c$.

\emph{Step 3 (error dynamics).} Inserting $\delta_{k+1}$ into the first line of Step~1 gives
\begin{equation*}
e_{k+1}=P_kM_ke_k+N_{A,k}(C_{k+1}N_{A,k})^{+}\eta_{k+1}+P_kG_k\omega_k+\tilde r_k,\qquad |\tilde r_k|\le\tilde\ell\,\big(|e_k|^2+|\eta_{k+1}|^2+|\omega_k|^2\big).
\end{equation*}
By (S3), the transition matrix $\Phi(k,j)$ of $P_kM_k$ satisfies $\|\Phi(k,j)\|\le c_0\rho_0^{\,k-j}$ with $\rho_0=\rho_L^{1/L}$ and $c_0$ the maximum norm of products of fewer than $L$ factors. Variation of constants gives
\begin{equation*}
|e_k|\le c_0\rho_0^{\,k}|e_0|+c_1\sum_{j=0}^{k-1}\rho_0^{\,k-1-j}\big(|\eta_{j+1}|+|\omega_j|\big)+c_0\tilde\ell\sum_{j=0}^{k-1}\rho_0^{\,k-1-j}\big(|e_j|^2+|\eta_{j+1}|^2+|\omega_j|^2\big).
\end{equation*}

\emph{Step 4 (absorbing the remainder).} On the neighborhood, $|e_j|,|\eta_{j+1}|,|\omega_j|\le\varepsilon'$ for some $\varepsilon'$ proportional to $\varepsilon$, so each quadratic term is bounded by $\varepsilon'$ times the corresponding linear term. Choosing $\varepsilon$ such that $c_0\tilde\ell\varepsilon'/(1-\rho_0)\le(1-\rho_0)/2$ and applying a discrete Gronwall argument by induction on $k$ gives \eqref{eq:det1} with $\hat\rho=(1+\rho_0)/2$ and a suitable $c$.
\end{proof}

\begin{remark}
If all $n_c$ candidates are treated as unknown inputs, i.e., without complementarity, then $\operatorname{rank}(C_{k+1}N_k)\le6<n_c$ and (S2) fails. This is the dynamic counterpart of Proposition~\ref{prop:nonid}.
\end{remark}

\subsection{Numerical verification of the assumptions}\label{sec:verify}
We evaluated Assumption~\ref{ass:det} at the 805 contact configurations. The smallest eigenvalue of $\sym(H)$ was 0.36, so (S1) holds, and the free dynamics were contractive with $\rho(M)\le0.988$. The matrix $C N_A$ had full column rank in all 805 configurations, with a median smallest singular value of 2.2~mm/N and a minimum of 0.004~mm/N, so (S2) holds with a large but finite input-recovery gain at a few configurations. For (S3), the pointwise spectral radius of $P_kM_k$ was at most 0.967 (median 0.81); single steps could expand the norm by up to 11$\times$ near changes of the active set, so we evaluated products along full trajectories. Products over 20 consecutive samples (1~s) had norm at most 0.32, and over 40 samples at most 0.033, on runs 1, 3, and 5. With all 13 candidates as unknown inputs, $\operatorname{rank}(CN)\le6$ in every configuration.

\section{Constrained Moving Horizon Estimation}\label{sec:mhe}
This section presents the estimator, its real-time implementation, and its stability.

\subsection{Problem formulation}
At sample $k$, the MHE estimates the configurations and contact forces over the window $j=k-N,\dots,k$ by solving
\begin{equation}\label{eq:mhe}
\begin{aligned}
\min_{\{q_j,\lambda_j\}}\ & \|q_{k-N}-\bar q\|^2_{W}+\sum_{j=k-N}^{k}\frac{\|y_j-h(q_j)\|^2}{\sigma_y^2}+\sum_{j=k-N}^{k-1}\frac{\|r(q_{j+1},q_j,\lambda_{j+1},T_j)\|^2}{\sigma_q^2}+\rho\sum_{j}\mathbf{1}^{\top}\lambda_j\\
\text{s.t.}\ & \lambda_j\ge0,\quad g(q_j)\ge0,\quad \lambda_{j,i}\,g_i(q_j)\le\epsilon,\quad |q_j|_\infty\le0.6,
\end{aligned}
\end{equation}
where $h(q)=l(q)-l(0)$; $r$ is the residual of \eqref{eq:dyn}, which eliminates the process noise as a decision variable; $\bar q$ and $W$ define the arrival cost; $\sigma_q$ weights the model residual; $\rho=0.05$ is a small weight that favors sparse forces; and $\epsilon=2\times10^{-4}$ relaxes complementarity. We chose the horizon $N=6$ (0.3~s) to be long enough to span the settling of the overdamped dynamics but short enough to keep the problem small. Solved to convergence with IPOPT \cite{wachter2006} through CasADi \cite{andersson2019}, using a Gauss--Newton Hessian of the Lagrangian (which gave the same solutions as the exact Hessian 2.7$\times$ faster), \eqref{eq:mhe} takes 580~ms per sample.

\subsection{Real-time iteration}
Following \cite{diehl2005,wynn2014}, we solve one Gauss--Newton quadratic program (QP) per sample, linearized at the window from the previous sample shifted by one step. The residual Jacobians come from small per-stage functions compiled to C, which compile in minutes, unlike a function for the whole window. We solve the QP with OSQP \cite{stellato2020}, warm-started and polished, with tolerances of $10^{-3}$.

Because complementarity is nonconvex, we replace it in the QP by a linearized penalty,
\begin{equation}\label{eq:pen}
\sum_{j,i}\big(\rho+\kappa\max(g_i(\bar q_j),0)\big)\,\lambda_{j,i},
\end{equation}
where $\bar q_j$ is the linearization point and $\kappa$ prices force at each candidate point by its current gap; we also fix $\lambda_{j,i}=0$ when $g_i(\bar q_j)>30$~mm. Non-penetration is kept as a linearized hard constraint. We first tried a hard rule that fixed $\lambda_{j,i}=0$ whenever $g_i(\bar q_j)>3$~mm. This rule failed: when the estimated robot drifted slightly away from the obstacle, the estimated force locked at zero for more than 1~s. The penalty \eqref{eq:pen} degrades gracefully instead, because a slightly misplaced contact point only raises the price of force there.

The real-time estimator takes 28~ms per sample on average (95th percentile 47~ms; 2.2\% of samples above the 50~ms budget) in Python on a two-core virtual machine, with accuracy equal to the fully converged MHE (Section~\ref{sec:results}).

\subsection{Stability}
\begin{theorem}[Local robust stability of the real-time MHE]\label{thm:stab}
Assume (S1)--(S3) hold for every visited active set, process and measurement disturbances are bounded by $\bar w$ and $\bar v$, and active sets persist for at least $\tau_d$ samples. Let $\varepsilon_p$ bound the linear perturbation of the QP cost introduced by \eqref{eq:pen}. Then there exist $\underline N$, $\tau^\ast$, $r_0>0$, $C\ge1$, $\eta\in(0,1)$, and gains $\gamma_w,\gamma_v,\gamma_p$ such that, for $N\ge\underline N$, $\tau_d\ge\tau^\ast$, and $|\hat x_0-x_0|\le r_0$, the real-time MHE satisfies, for all $k$,
\begin{equation}
|\hat x_k-x_k|\le\max\big\{C\eta^k|\hat x_0-x_0|,\ \gamma_w\bar w+\gamma_v\bar v+\gamma_p\varepsilon_p\big\},
\end{equation}
where $x=(q,\lambda)$.
\end{theorem}
\begin{proof}[Proof sketch]
\emph{Fixed active set, exact MHE.} Proposition~\ref{prop:det} gives local exponential strong detectability. Under this property, MHE with a sufficiently long horizon is a robust unknown-input estimator with exponential bounds \cite[Thm.~3.6]{guo2025}, which builds on the Lyapunov-based robust stability analysis of \cite{schiller2023} and the detectability framework of \cite{allan2021}.

\emph{Real-time iteration.} With full-rank Gauss--Newton Jacobians, the single-QP scheme inherits the contraction of the exact MHE in a neighborhood of the solution, up to an additional term that is quadratic in the linearization error \cite{wynn2014}. Choosing $r_0$ small keeps this term dominated by the contraction.

\emph{Penalty.} The QP has a positive-definite Gauss--Newton Hessian on the free directions and linearly independent active constraints, so its solution is unique and Lipschitz in the linear cost term. The penalty \eqref{eq:pen} therefore shifts the estimate by at most $\gamma_p\varepsilon_p$.

\emph{Changes of the active set.} The Lyapunov functions of adjacent active sets are equivalent on the neighborhood, so the error bound grows by at most a factor $\mu\ge1$ at each change. If $\mu\,\eta^{\tau_d}<1$, i.e., $\tau_d\ge\tau^\ast=\ln\mu/\ln(1/\eta)$, a dwell-time argument \cite{liberzon2003} preserves exponential decrease across changes.
\end{proof}

\begin{remark}[Scope]
Theorem~\ref{thm:stab} is local and mode-wise. Its detectability assumptions are proven to imply the needed property (Proposition~\ref{prop:det}) and are verified numerically along the simulated trajectories (Section~\ref{sec:verify}), but the constants are not computed. Section~\ref{sec:conv} checks the predicted behavior empirically.
\end{remark}

\section{Simulation Study}\label{sec:results}
This section compares the estimators on a finite-element ground truth. We first describe the setup and baselines, and then report the nominal comparison, robustness tests, convergence, model mismatch, and a control experiment without contact.

\subsection{Ground truth}
We simulated the robot in SOFA \cite{faure2012} with a corotational Timoshenko beam of 24 elements on rigid frames, with physical disk mass and rotational inertia. Routed tendons apply their load as distributed local moments $T_i\,r\,(-\sin\alpha_i,\cos\alpha_i,0)$ in the frame of each element they pass through. We chose this representation because SOFA's corotational beam mishandled the large follower compression that the equivalent point forces produce, while the axial strain neglected in this way is below 0.5\%. The wall is a stiff penalty contact (3000~N/m) on the backbone points, offset by the disk radius. Five 40~s runs with randomized commands placed the wall 0.26--0.31~m from the base axis; the robot was in contact 67--94\% of the time, with forces up to 4.7~N at 0.525--0.6~m of arc length. We tuned all estimators on a separate run.

\subsection{Baselines}
We compare against four estimators. PCC inversion fits a piecewise-constant-curvature model \cite{webster2010} to the tendon lengths by least squares. The other three share the model \eqref{eq:dyn}--\eqref{eq:y}: an EKF without contact; an EKF with the 13 contact forces as random-walk states; and the same EKF with projection of each estimate onto $\{\lambda\ge0,\ g(q)\ge0\}$ by a QP \cite{simon2010}. We tuned the EKFs' force process noise on the tuning run over a grid of 0.02--0.5~N. The best values were 0.05~N without and 0.1~N with projection; an earlier untuned value of 0.5~N had overstated the MHE's force advantage by approximately 2.5$\times$.

\subsection{Nominal comparison}
\begin{figure}[t]
\centering
\includegraphics[width=0.62\textwidth]{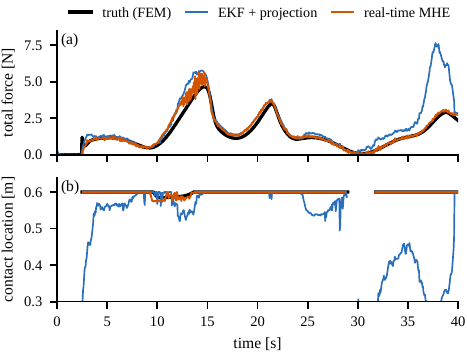}
\caption{Contact estimation over time (run~1). (a) Total contact force. (b) Force-weighted contact location along the arc length, shown when the force exceeds 0.3~N.}
\label{fig:time}
\end{figure}

The real-time MHE estimated total contact force with an RMSE of 0.22~N and contact location with an error of 0.9~mm, compared with 0.72~N and 65~mm for the projected EKF (Table~\ref{tab:main}). Shape error was the same for both (3.0 and 3.2~mm), which shows that the gain comes from the contact constraints rather than from the model. Without projection, the EKF placed 6.6~N of force at points that were not in contact and estimated the robot 25.8~mm inside the wall, as Proposition~\ref{prop:nonid} predicts. Figure~\ref{fig:time} shows the effect over time: the projected EKF overestimated the force at 38~s (7.6~N versus 2.9~N) and its location estimate moved between candidate points by up to 0.3~m, while the MHE remained at the true contact point. The real-time MHE matched the fully converged MHE, with a slightly lower force error, likely due to the smoothing effect of the penalty \eqref{eq:pen}.

\begin{table}[t]
\centering
\caption{Mean (standard deviation) over five contact runs. Location: error in the force-weighted contact arc length. Spurious: mean force estimated at points more than 10~mm from the wall. Penetr.: maximum estimated penetration into the wall. Detect.: fraction of samples with correct contact detection (threshold 0.3~N).}
\label{tab:main}
\small
\begin{tabular}{@{}lccccccc@{}}
\toprule
 & Shape & Force & Location & Spurious & Penetr. & Detect. & Time \\
Estimator & (mm) & (N) & (mm) & (N) & (mm) & (\%) & (ms) \\
\midrule
PCC & 13.8 (2.3) & -- & -- & -- & -- & -- & 0.5 \\
EKF, no contact & 23.0 (3.2) & -- & -- & -- & -- & -- & 91 \\
EKF + forces & 5.6 (0.4) & 2.44 (0.62) & -- & 6.6 & 25.8 & 41 & 137 \\
EKF + projection & 3.2 (0.3) & 0.72 (0.26) & 65 (12) & 0.50 & 9.0 & 95 & 140 \\
MHE, IPOPT & 3.0 (0.3) & 0.25 (0.12) & 2.1 (0.6) & 0.009 & 0 & 98 & 580 \\
\textbf{MHE, real time} & 3.0 (0.3) & \textbf{0.22} (0.08) & \textbf{0.9} (0.8) & $<$0.001 & 0.04 & 98 & \textbf{28} \\
\bottomrule
\end{tabular}
\end{table}

\begin{figure}[t]
\centering
\includegraphics[width=\textwidth]{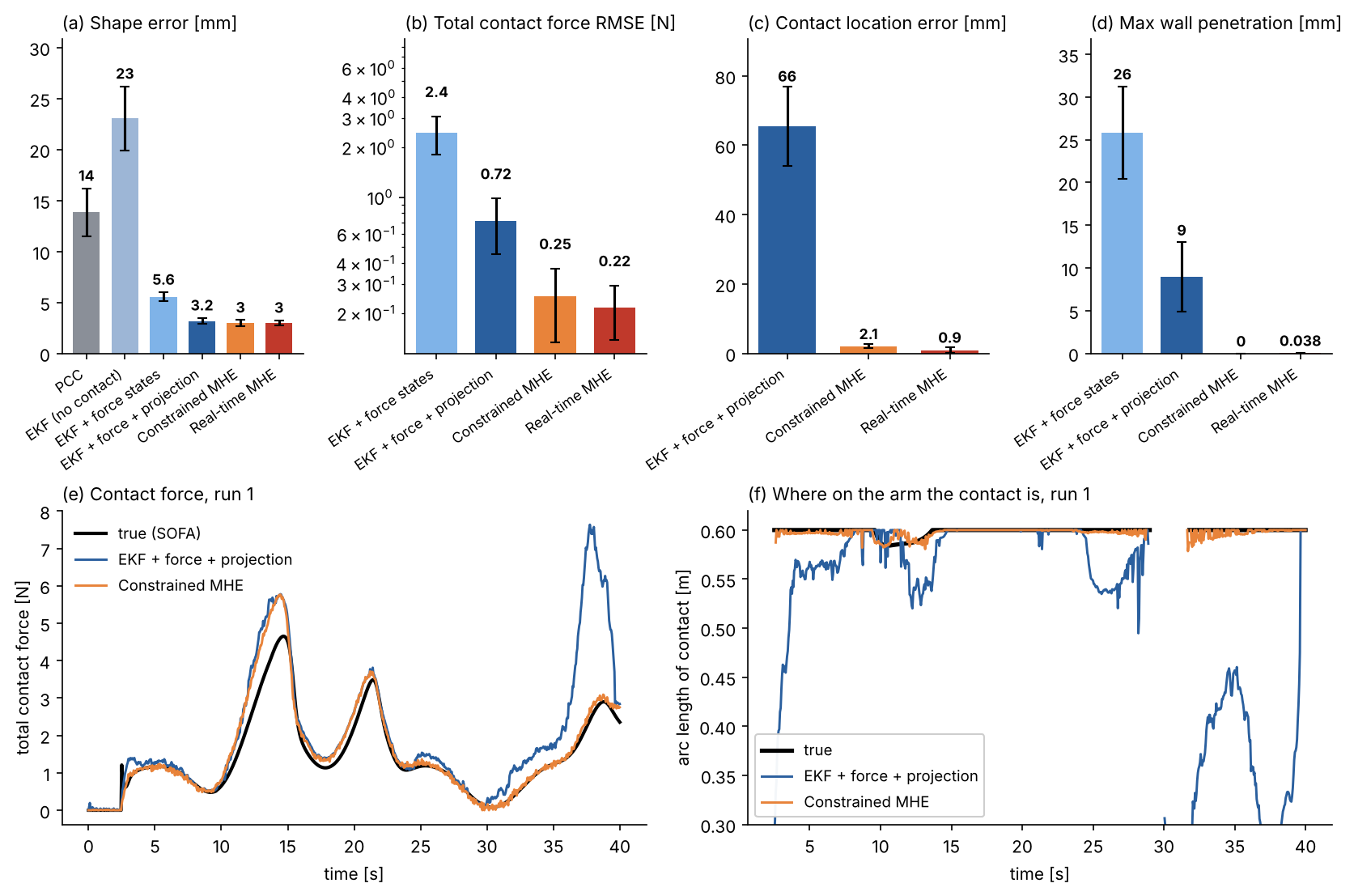}
\caption{Nominal comparison over five contact runs. Top: shape error, total contact force RMSE (log scale), contact location error, and maximum wall penetration for each estimator. Bottom: total contact force and contact location over time for run~1.}
\label{fig:summary}
\end{figure}

\subsection{Robustness}\label{sec:rob}
\begin{figure}[t]
\centering
\includegraphics[width=\textwidth]{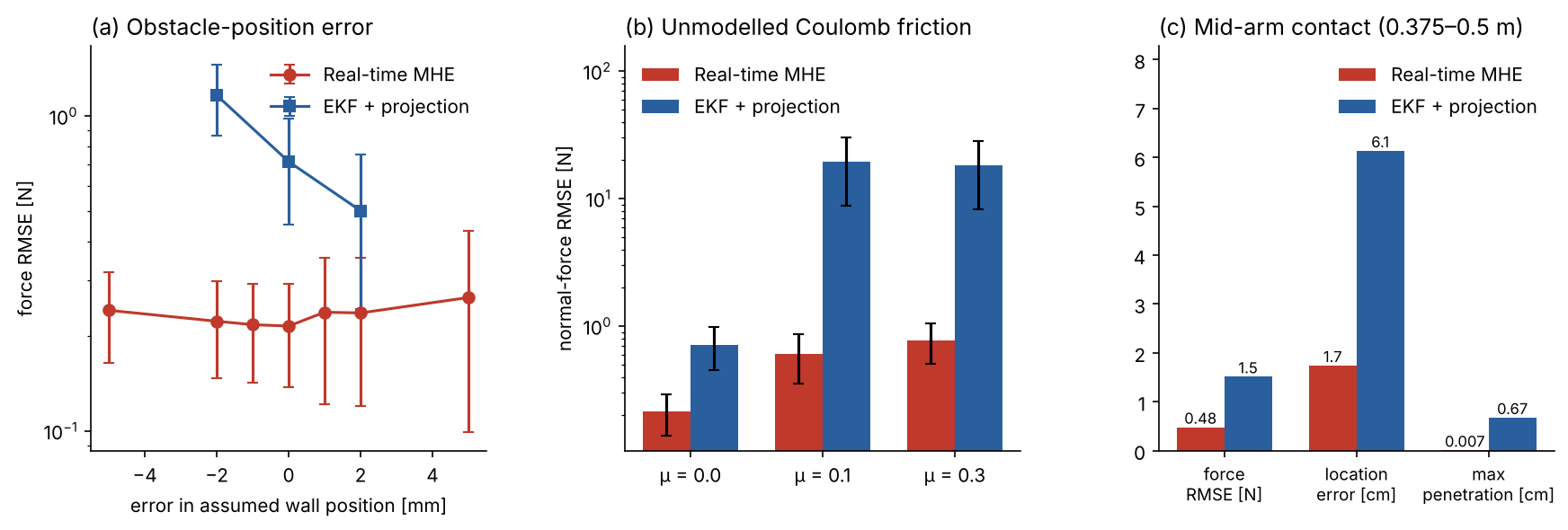}
\caption{Robustness. (a) Force RMSE versus error in the assumed wall position. (b) Normal-force RMSE under unmodeled Coulomb friction at the wall. (c) Mid-body contact.}
\label{fig:rob}
\end{figure}

Table~\ref{tab:rob} and Fig.~\ref{fig:rob} summarize four further tests with unchanged tuning. First, an error of $\pm5$~mm in the assumed wall position raised the MHE's force error from 0.22 to at most 0.27~N, and its reported penetration, measured against its own wrong wall, stayed below 0.7~mm. The projected EKF's location error nearly doubled when the wall was assumed 2~mm too close. Second, with Coulomb friction at the wall ($\mu=0.1$ and $0.3$), implemented in the ground truth as tangential springs with slip at up to 2~N and modeled by no estimator, the MHE's force error stayed below 0.8~N and its location error near 1~mm. The projected EKF showed repeated excursions to 30--150~N, likely because small tangential loads push its estimate along $\ker G$. Both estimators' shape errors rose above 22~mm, because friction adds lateral loads and torsion that the model cannot explain. Third, with contact near the middle of the robot (0.375--0.5~m, with the wall at 0.13--0.16~m), the MHE's location error rose to 17~mm, because contact spread across several candidate points spaced 50~mm apart; its force error of 0.48~N remained 3$\times$ lower than the EKF's.

\begin{table}[t]
\centering
\caption{Robustness with unchanged tuning (five runs per condition): force RMSE in N, location and shape error in mm.}
\label{tab:rob}
\small
\begin{tabular}{@{}lcccccc@{}}
\toprule
 & \multicolumn{3}{c}{Real-time MHE} & \multicolumn{3}{c}{EKF + projection} \\
\cmidrule(lr){2-4}\cmidrule(l){5-7}
Condition & Force & Location & Shape & Force & Location & Shape \\
\midrule
Nominal & 0.22 & 0.9 & 3.0 & 0.72 & 65 & 3.2 \\
Wall assumed 2~mm too close & 0.22 & 1.0 & 3.1 & 1.16 & 118 & 3.5 \\
Wall assumed 2~mm too far & 0.24 & 0.9 & 3.2 & 0.50 & 36 & 3.2 \\
Wall assumed $\pm5$~mm off & $\le$0.27 & $\le$1.0 & $\le$3.8 & -- & -- & -- \\
Wall friction $\mu=0.1$ & 0.61 & 1.3 & 22.4 & 19.4 & 286 & 28.7 \\
Wall friction $\mu=0.3$ & 0.78 & 1.4 & 23.1 & 18.2 & 269 & 29.0 \\
Mid-body contact & 0.48 & 17 & 5.0 & 1.51 & 61 & 5.9 \\
\bottomrule
\end{tabular}
\end{table}

\subsection{Convergence}\label{sec:conv}
\begin{figure}[t]
\centering
\includegraphics[width=\textwidth]{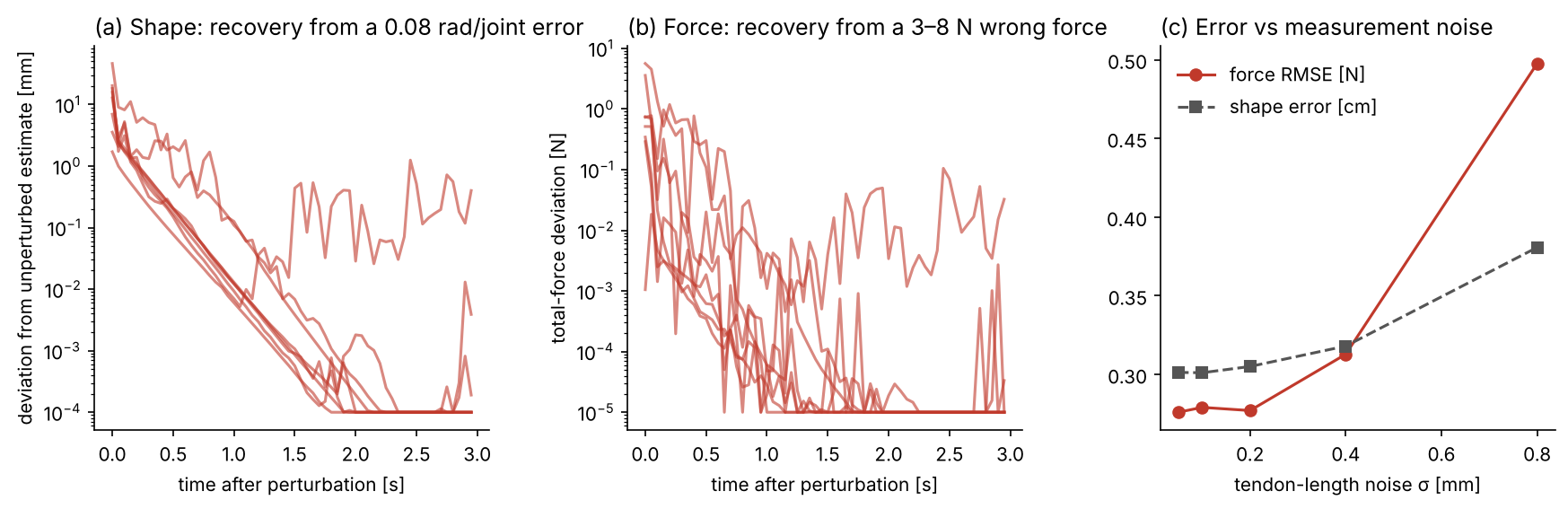}
\caption{Convergence of the real-time MHE. (a) Shape and (b) total-force deviation from the unperturbed estimate after eight large perturbations of the estimation window. (c) Force RMSE and shape error versus tendon-length noise.}
\label{fig:conv}
\end{figure}

To test the behavior predicted by Theorem~\ref{thm:stab}, we perturbed the whole estimation window eight times during run~1, by 0.08~rad per joint and by a wrong force of 3--8~N at a random candidate point, giving initial deviations from the unperturbed estimate of up to 46~mm and 5.6~N. In all eight cases, the estimate returned to within 0.03~mm and 0.004~N of the unperturbed estimate within 2~s, decaying by approximately 10$\times$ every 0.25~s (Fig.~\ref{fig:conv}(a,b)). We also scaled the tendon-length noise from 0.05 to 0.8~mm. The force RMSE stayed at 0.28~N up to 0.2~mm and rose to 0.31~N at 0.4~mm and 0.50~N at 0.8~mm (Fig.~\ref{fig:conv}(c)): the error is set by model mismatch until noise dominates, after which it grows with a bounded gain, as the theorem predicts.

\subsection{Model mismatch}
\begin{figure}[t]
\centering
\includegraphics[width=\textwidth]{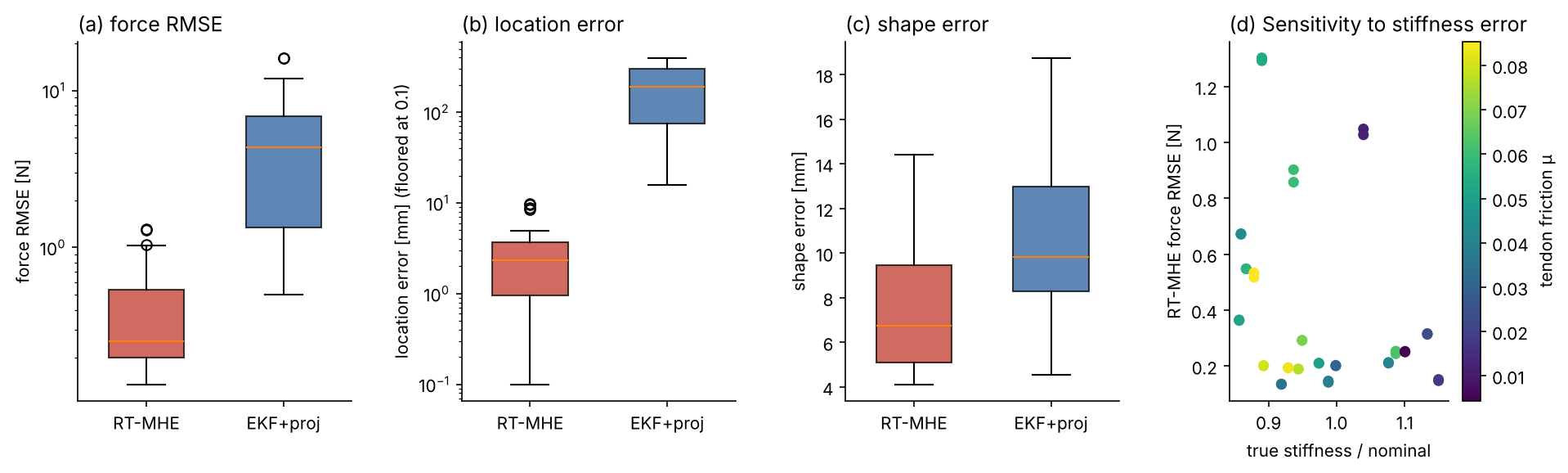}
\caption{Model-mismatch Monte Carlo (20 runs, two noise realizations each). (a) Force RMSE. (b) Contact location error (values below 0.1~mm shown at 0.1~mm). (c) Shape error. (d) Real-time MHE force RMSE versus true stiffness, colored by tendon friction.}
\label{fig:mc}
\end{figure}

To test the estimator against a robot that differs from its model, we simulated 20 further runs in which the true stiffness was scaled by 0.86--1.15, the tendon offset by 0.94--1.07, and the structural damping by 0.72--1.28, and in which the tendon tension decayed along its routing with a capstan friction coefficient of 0--0.085. None of these effects is modeled. We evaluated each run with two noise realizations (40 trials) and unchanged tuning. Table~\ref{tab:mc} and Fig.~\ref{fig:mc} report the results. The MHE was better in force and location in all 40 trials (Wilcoxon signed-rank test, $p=2\times10^{-12}$). Mismatch roughly doubled the MHE's force error (0.22 to 0.40~N; 95th percentile 1.06~N) and raised its shape error from 3.0 to 7.5~mm, while the projected EKF degraded by 7$\times$ in force and 3$\times$ in location. The largest MHE force errors occurred when a soft robot (stiffness scaled by 0.86--0.9) was combined with tendon friction.

\begin{table}[t]
\centering
\caption{Model-mismatch Monte Carlo (40 trials): mean and 95\% bootstrap confidence interval.}
\label{tab:mc}
\small
\begin{tabular}{@{}lcccc@{}}
\toprule
Metric & Real-time MHE & EKF + projection & MHE better & Wilcoxon $p$ \\
\midrule
Force RMSE (N) & 0.40 [0.31, 0.51] & 4.8 [3.7, 6.1] & 40/40 & $2\times10^{-12}$ \\
Location error (mm) & 2.8 [2.0, 3.6] & 193 [155, 232] & 40/40 & $2\times10^{-12}$ \\
Shape error (mm) & 7.5 [6.6, 8.4] & 10.7 [9.6, 12.0] & 30/40 & $7\times10^{-6}$ \\
Max penetration (mm) & 0.37 [0.15, 0.65] & 15.6 [10.4, 22.1] & 37/40 & $3\times10^{-11}$ \\
Contact detection (\%) & 94 & 83 & -- & -- \\
\bottomrule
\end{tabular}
\end{table}

\subsection{Control experiment: motion without contact}
To test whether the MHE's advantage depends on the contact constraints, we ran a separate study without obstacles, in which an unknown payload of 40--120~g (nominal) or 150--350~g (heavy) was attached to the tip during each 40~s run, and in which 2\% of the length measurements were replaced by 3--8~mm outliers (outliers). The estimators used the model without torsion and estimated the payload as an additional state, constrained to be non-negative in the MHE. Table~\ref{tab:free} reports the results over five runs per condition. The MHE and the EKF reached the same shape and payload accuracy in every condition; a robust (pseudo-Huber) MHE reduced shape error under outliers by 9\%, which is within one standard deviation. Both model-based estimators were limited by the reduced model: with the true payload and no noise, the model alone gave shape errors of 4--13~mm. Without contact constraints, the MHE therefore offers no advantage, which supports our conclusion that its benefit comes from encoding complementarity.

\begin{table}[t]
\centering
\caption{Motion without contact (five runs per condition): mean backbone shape error in mm (standard deviation) and payload RMSE in g.}
\label{tab:free}
\small
\begin{tabular}{@{}lcccccc@{}}
\toprule
 & \multicolumn{2}{c}{Nominal} & \multicolumn{2}{c}{Heavy payload} & \multicolumn{2}{c}{Outliers} \\
\cmidrule(lr){2-3}\cmidrule(lr){4-5}\cmidrule(l){6-7}
Estimator & Shape & Payload & Shape & Payload & Shape & Payload \\
\midrule
PCC & 4.14 (0.39) & -- & 9.42 (1.46) & -- & 5.24 (0.40) & -- \\
EKF & 3.52 (0.37) & 8.3 & 7.70 (2.08) & 20.4 & 4.01 (0.37) & 10.2 \\
MHE & 3.55 (0.38) & 8.3 & 7.75 (2.10) & 20.1 & 3.98 (0.39) & 10.8 \\
Robust MHE & 3.55 (0.39) & 8.6 & 7.89 (2.09) & 22.4 & 3.65 (0.37) & 9.0 \\
\bottomrule
\end{tabular}
\end{table}

\section{Discussion and Conclusion}\label{sec:concl}
We presented an analysis and an estimator for locating contact on tendon-driven continuum robots using only tendon lengths and tensions. A key takeaway is that the information needed to locate contact is not in the measurements but in the geometry: tendon measurements leave at least seven force directions unobservable, and a known obstacle with complementarity restores identifiability and strong detectability. Using this insight, we presented a constrained MHE that runs in real time and locates contact to approximately 1~mm in a finite-element benchmark, where a tuned EKF with projection does not. The control experiment without contact, in which the MHE and the EKF performed identically, confirms that the benefit comes from the constraints rather than from the optimization.

The approach is well suited to tasks in which the geometry of the surroundings is known or can be measured beforehand, such as instrument guidance against registered anatomy or manipulation inside known enclosures. There are situations in which it would not be the choice, such as contact with unknown or moving obstacles, where shape sensing remains necessary. One question that remains open is how best to combine the two: for example, by estimating a parametrized obstacle together with the contact forces.

This work has several limitations, at least in its current form. First, all results are in simulation; the ground truth is a beam finite-element model, and effects such as hysteresis, manufacturing tolerances, and sensor bias remain untested. Second, the estimator assumes a known planar obstacle and frictionless contact; friction preserved force accuracy but degraded shape estimation to more than 22~mm. Third, contact near the middle of the robot spreads over adjacent candidate points, which raised the location error to 17~mm; a finer or continuous contact parametrization would likely reduce it. Fourth, Theorem~\ref{thm:stab} is local and mode-wise, with constants that we did not compute, and the conditions of Proposition~\ref{prop:det} were verified numerically along the simulated trajectories rather than proven for all configurations. Finally, timing was measured in Python, and a compiled implementation is needed for hard real-time guarantees. Future work will validate the estimator on a tendon-driven prototype pressing on an instrumented surface, add friction cones and curved obstacles to the constraints, and quantify how uncertainty in the obstacle geometry propagates to the contact estimate.

\section*{Acknowledgment}
Portions of the manuscript text, the simulation and analysis code, and the figures were drafted with the assistance of the AI system Claude (Anthropic) \cite{claude2026}; the author reviewed, verified, and takes full responsibility for all content.

\section*{Data and code availability}
The simulation scenes, estimator code, and result data are available from the author upon request.

\bibliographystyle{IEEEtran}
\bibliography{refs}

\end{document}